\documentclass[letterpaper, 10pt, conference]{ieeeconf}  
\pdfoutput=1

\IEEEoverridecommandlockouts                              

\usepackage{xcolor}
\usepackage{amsmath,bm,times}
\usepackage{graphicx}
\usepackage{cite}
\usepackage{url}
\usepackage{amsmath,amssymb,amsfonts}
\usepackage{algorithmic}
\usepackage{graphicx}
\usepackage{textcomp}

\usepackage{subcaption}

\usepackage{multirow}

\usepackage{hyperref}

\newcommand{\OUT}[1]{}

\title{\LARGE \bf
An End-to-End Human Simulator for Task-Oriented Multimodal Human-Robot Collaboration}

\author{Afagh Mehri Shervedani$^{1}$, Siyu Li$^{1}$, Natawut Monaikul$^{2}$, Bahareh Abbasi$^{3}$, Barbara Di Eugenio$^{2}$,\\and Milo\v s \v Zefran$^{1}$
\thanks{$^{1}$A.M. Shervedani, S. Li, and Milo\v s \v Zefran are with the Robotics Laboratory, Department of Electrical and Computer Engineering, University of Illinois Chicago, Chicago, IL 60607 USA.}%
\thanks{$^{2}$N. Monaikul and B. Di Eugenio are with the Natural Language Processing Laboratory, Department of Computer Science, University of Illinois Chicago, Chicago, IL 60607 USA.}%
\thanks{$^{3}$B. Abbasi is with the Computer Science Department, California State University Channel Islands, Camarillo, CA 93012 USA.}%
\thanks{ This work has been supported by the National Science Foundation grants IIS-1705058 and CMMI-1762924.}%
}

\begin{document}

\maketitle

\begin{abstract}
This paper proposes a neural network-based user simulator that can provide a multimodal interactive environment for training Reinforcement Learning (RL) agents in collaborative tasks involving multiple modes of communication. The simulator is trained on the existing ELDERLY-AT-HOME corpus and accommodates multiple modalities such as language, pointing gestures, and haptic-ostensive actions. The paper also presents a novel multimodal data augmentation approach, which addresses the challenge of using a limited dataset due to the expensive and time-consuming nature of collecting human demonstrations. Overall, the study highlights the potential for using RL and multimodal user simulators in developing and improving domestic assistive robots.
\end{abstract}

\section{Introduction}
\label{Intro}

\begin{figure}[t]
\vspace{2mm}
\centering
\includegraphics[width=\columnwidth]{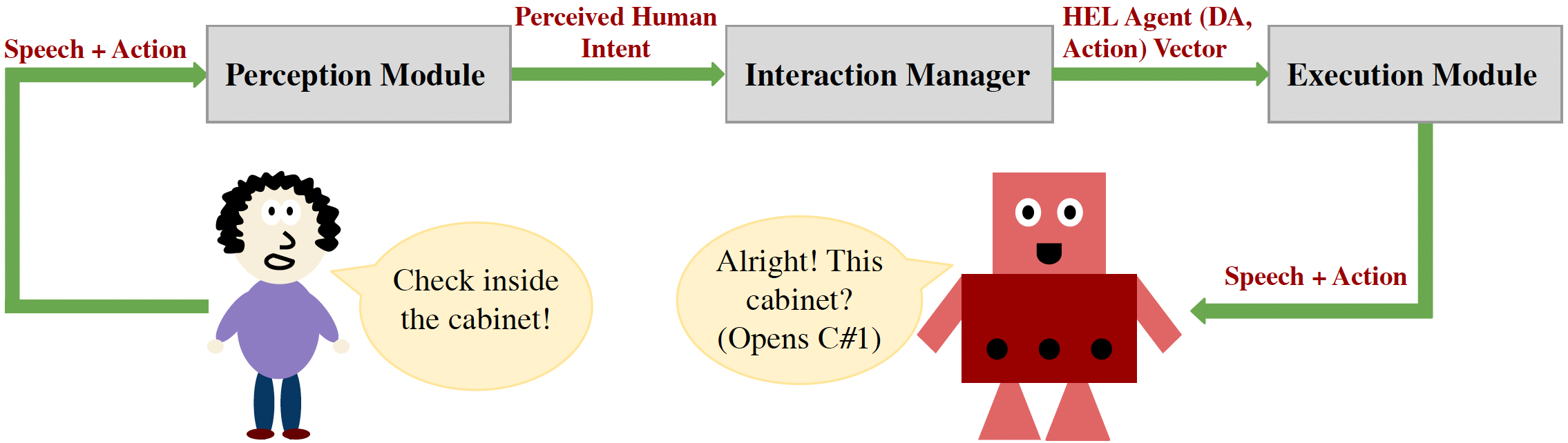}
\caption{The \emph{Sense-Plan-Act} cycle in an assistive robot. See also \cite{shervedani2023multimodal} for a more detailed illustration. 
}
\label{fig:user_interaction}
\end{figure}

Assistive robots designed for domestic use can be instrumental in enabling older adults and those with disabilities to maintain their independence by providing assistance with activities of daily living (ADLs) such as cooking and cleaning. These robots typically rely on three primary modules to perform the sense-plan-act cycle: the \emph{Perception Module} processes sensory information, the \emph{Execution Module} carries out desired actions, and perhaps the core of the robot, the \emph{Interaction Manager} that receives the input from the Perception Module and determines the most effective course of action for the robot to take. Figure \ref{fig:user_interaction} illustrates the sense-plan-act cycle.

For a robot to assist in an ADL as a human would, it must be able to both effectively collaborate on the task as well as engage naturally in interaction with its user. This interaction is typically multimodal in nature, seamlessly interleaving various modes of communication, such as language, gestures, facial expressions, and force exchanges. An Interaction Manager designed for such a human-robot collaboration must therefore be robust enough to handle these multiple modalities.



In previous work~\cite{abbasi2019multimodal,monaikul2020role}, we developed a novel framework named Hierarchical Bipartite Action-Transition Networks (HBATNs) as a basis for an \emph{Interaction Manager} for assistive robots participating in what we call the \textit{Find} task. In this task, two participants, one of whom may have limited mobility, collaborate to locate an object. The design of the HBATNs draws directly from human demonstrations in our ELDERLY-AT-HOME corpus~\cite{chen2015roles}, which contains annotated interactions between nursing students (labeled HEL) and elderly individuals (labeled ELD) receiving assistance with ADLs. Although the HBATNs provide an optimal policy for a robot participating in the ``Find'' task based on human demonstrations, they were manually constructed and can not be easily adapted to other tasks.



In our recent study~\cite{shervedani2023multimodal}, we proposed a more scalable \emph{Interaction Manager} that employs reinforcement learning (RL) to automatically extract an optimal policy for a robot participating in a collaborative task. One significant challenge in training this agent is building an interactive environment that can provide the agent with rewards. In task-oriented collaborative tasks, the interactive environment involves the human. Since having a human interact with an Artificial Intelligence (AI) agent during the RL training phase is not practical, researchers have developed user simulators to simulate human behavior in such interactions. However, existing user simulators only support a single modality, which is language. Therefore, we developed our own user simulator to imitate human behavior in multimodal interactions. The simulator needs to interpret a multimodal input and generate multimodal responses.

In this paper, we propose a novel neural network-based user simulator that is inspired by Behavioral Cloning ~\cite{bratko1995behavioural, torabi2018behavioral} and is trained on the ``Find'' task demonstrations in the ELDERLY-AT-HOME corpus. What makes our user simulator unique is the fact that it accommodates multiple modalities such as language, pointing gestures, and haptic-ostensive (H-O) actions~\cite{chen2015roles}. This is a substantial advance in developing domestic assistive robots as there is currently no simulator that can provide a multimodal interactive environment for RL training.

Another significant contribution is our novel multimodal data augmentation approach, which addresses the challenge of using a limited dataset due to the expensive and time-consuming nature of collecting human demonstrations. Our data augmentation approach effectively takes care of all modalities involved in the data, making it a valuable resource for the community. Overall, our study highlights the potential for using RL and multimodal user simulators in developing and improving domestic assistive robots, and represents a significant development for the field.

The rest of the paper is structured as follows. First, we provide a review of related studies in Section~\ref{sec: Related Work}. We provide the necessary preliminaries in Section~\ref{MIM}. Our user simulator framework is presented in Section~\ref{Framework}, and the evaluations are described in Section~\ref{Evaluation}. Finally, we summarize our findings and provide concluding remarks in Section~\ref{conclusion}.


\section{Related Work}
\label{sec: Related Work}

Simulators are an important tool in developing robot autonomy through machine learning. The challenge for human-robot interaction (HRI) and the development of assistive robots is that the simulators need to model complex human behavior. Recent work in HRI and human factors has been successful in describing, understanding, and predicting human decision-making in specific contexts~\cite{goodrich2008human}. The approaches relevant to HRI can be broadly divided into three categories.
One common technique for modeling human behavior involves modeling the mechanics of human movement through equations of motion~\cite{peternel_towards_2017, erickson2020assistive}.
Another technique assumes that the human agent has a reward function nearly identical to that of the robot. This assumption is frequently made in the human-robot interaction community~\cite{nikolaidis2017mathematical}. For example, in a scenario where the human and robot agents pick a tool from a table, the reward is defined as a function over the $xy$-plane. What's more, many works on human-robot interactions assume that the human agent is an expert in interacting with the robot \cite{wu2022survey, sharkawy2021human}, and thus, humans will always choose actions that are easy for the robot to understand. Set an example of  
 ~\cite{sharkawy2021human}, the human agent's policy is a fixed and deterministic function of the robot's actions. However, all these techniques impose strong assumptions on human agents. In our work, we relax some of these assumptions and model the human agents based on real-world data.

In the realm of HRI, much attention is on care for older adults in the home setting. Our previous work~\cite{abbasi2019multimodal, monaikul2020role} uses HBATNs to model both the robot helper and the older person simultaneously to maintain the state of task-driven multimodal interaction and plan subsequent moves. The behavior of the human has been assumed to consist of dialogue actions, pointing gestures, and haptic-ostensive cues. Similar works in the field, such as~\cite{ranieri2018lara}, have introduced a multimodal action recognition framework that processes speech and video information separately and then fuses them. In~\cite{chao2012timing},  a timed Petri net is proposed to represent socially intelligent robot behaviors, emphasizing the importance of the representation's ability to model time and asynchronous events. However, these works have been evaluated using hand-built human simulators, making it difficult to generalize to other tasks. In this paper, we build upon the HBATN framework by providing a more general version of the user simulator that can perform multimodality in the interactive environment.


As machine learning rapidly advances, new opportunities for human-robot interaction (HRI) arise. Recent advancements in intelligent dialogue systems have shown promise in modeling the dialogue aspect of human agents \cite{el2016sequence, kreyssig2018neural, li2017end}. For example, Li \cite{li2017end} developed a hybrid user simulator capable of naturally interacting with a dialogue system to perform end-to-end training. However, when considering scenarios where humans can perform more than just dialogue, user simulator research becomes more complex. \cite{choudhury2019utility} identified three ways to model how human changes affect the robot state: treating humans as part of the environment, having the robot learn how humans react, or assuming humans have agency and their own "theory of mind." Although the first two ways provide easier black-box solutions, our approach that falls into the third category allows for a more nuanced understanding of human behavior and intentions. However, training such models is challenging due to limited available data compared to traditional machine learning. To address this, we propose a human simulator that has shown promise in dealing with small datasets.
\section{Preliminaries}
\label{MIM}
Our previous studies~\cite{abbasi2019multimodal, monaikul2020role, shervedani2023multimodal} were developed based on the ELDERLY-AT-HOME corpus~\cite{chen2015roles}, a publicly available corpus of human-human multimodal interactions. The corpus involves performing assisted ADLs, such as putting on shoes and preparing dinner. We developed the framework of our HBATN and later our RL framework on a subcorpus consisting of the interactions related to the \emph{Find} task. That is, the elderly participant (ELD) would ask for an object, and the helper (HEL) would try to find it by asking follow-up questions.


In~\cite{abbasi2019multimodal}, the \emph{Find} task is decomposed into \OUT{as} a set of subtasks \OUT{with the goal of}to identify two main unknowns: the target object ($O$) and its location ($L$)\OUT{ of the object}. \OUT{As a result, we modeled }The four main subtasks are determining the desired object type (\textit{Det($O_{T}$)}), determining a potential location to check (\textit{Det($L$)}), opening the location (\textit{Open($L$)}), and determining the actual object (\textit{Det($O$)}). These are modeled as Action-Transition Networks (ActNets).

The ActNet is a bipartite graph representing the states of both participants \OUT{(ELD and HEL)}and their possible multimodal actions, which are defined as vectors consisting of linguistic features (the \emph{dialogue act} (DA)~\cite{chen2015roles} of the utterance, i.e. the speaker's intent and object or location words) and physical features (pointing gestures or \emph{haptic-ostensive} (H-O) actions). The HBATN encompass\OUT{ing} these ActNets \OUT{then }allowing a robot to not only infer the state of its partner but also to plan its next action accordingly.

Subsequently, in~\cite{monaikul2020role}, we generalized our model\OUT{ even more} to enable the robot to be either the ELD or HEL by decomposing \OUT{each subtask into finer subtasks, which we call \emph{primitive subtasks}}the subtasks into what we call \emph{primitive subtasks}. In this new formulation, \emph{Det($O_T$)} and \emph{Det($L$)} \OUT{consist of } establish the object type \OUT{or}and its location (\emph{Estab}), potentially followed up by verification (\emph{Verify}) or \OUT{follow-up}questions specifying for more information (\emph{Spec}), and\OUT{ that} \emph{Det($O$)} \OUT{consists of }confirms the presence or absence of the desired object (\emph{Finish}) in the current location or verify a physical object with the partner. We showed that our HBATN, equipped with a trained classifier that determines which subtask the participants are currently in, can model and perform both HEL and ELD behavior.

Finally, in our most recent study~\cite{shervedani2023multimodal}, we proposed an RL approach to extract the optimal policy in human-robot interaction. This policy replaces the hand-crafted HBATN policy for our HEL robot. As described in section \ref{Intro}, a major challenge throughout the RL training process is the interactive environment in which the agent needs to be placed. As a result, we developed a basic user simulator to act as the ELD and interact with the RL agent during the training.

The user simulator we developed in~\cite{shervedani2023multimodal}, which we call the \textbf{\emph{Basic User Simulator Model (BUS Model)}} was good enough to provide the right intent to the HEL agent. However, improving such a user simulator is still a challenging research question. The rest of this paper is focused on our approach to making the user simulator more accurate. We call our improved model the \textbf{\emph{Generic User Simulator Model (GUS Model)}}.

\section{User Simulator Framework}
\label{Framework}

\subsection{Feature Extraction}
\label{Feature Extraction}
In this work, we extend our previous works and introduce an end-to-end model to predict the state of the world that includes the ELD state as well as the ELD’s next move.


Our model is implicitly supposed to act as the ELD. However, our analysis of the data revealed that HEL could take consecutive moves. In order to address those cases where the HEL is taking more than one move, the model should determine whether or not the ELD will take an action in the subsequent move. In summary, the model's objective is twofold: (1) To predict the ELD's next action and whether it will occur. The ELD's move is determined by the ELD's DA and the ELD's action. (2) To determine the state of the world. The state of the world is determined by the ELD’s belief of the HEL’s knowledge of object type, location, and object.


In order to effectively train our end-to-end model, we must first determine what information can aid the model to decide the ELD's state and its action. The following items summarize the important points that should be taken into account when selecting the features and how we have featurized and annotated our collected human-human data.


\begin{itemize}
    \item 
    For determining whether the HEL is going to take consecutive moves or not, we need to know the previous actor (the current actor is the HEL). This information can be represented as a vector of length two. If the trial has not been started yet and the current actor, HEL, is initiating the trial, both elements are 0. If the HEL is not going to take consecutive moves, the previous actor is ELD and the first element of the vector is 1. Otherwise, the second element is 1.
    

    
   \item It is important to know whether or not an object type or location has been uttered. These can be represented by binary features.
    

    \item  
  The model also needs to have information on the ELD's previous state. We propose the following  representation for it: (1) ELD's belief of HEL's knowledge of $O_T$, which can be one of three values (ELD believes HEL does not know the target $O_T \rightarrow 0$, ELD believes HEL knows the target $O_T \rightarrow 1$, or ELD believes HEL is thinking of a different $O_T \rightarrow 2$), (2) ELD's belief of HEL's knowledge of $L$, and (3) ELD's belief of HEL's knowledge of $O$. We extracted all the meaningful possible combinations of these three parameters which would give us 13 distinct combinations as follows: (1) state $(0,0,0)$, (2) state $(0,1,0)$, (3) state $(0,2,0)$, (4) state $(1,0,0)$, (5) state $(1,0,1)$, (6) state $(1,0,2)$, (7) state $(1,1,0)$, (8) state $(1,1,1)$, (9) state $(1,1,2)$, (10) state $(1,2,0)$, (11) state $(2,0,0)$, (12) state $(2,1,0)$, (13) state $(2,2,0)$. We use one-hot encoding, a one-hot encoded vector of size thirteen, for representing these states. The elaborated description for obtaining these states is provided in section \ref{HBATN_evaluation}.
    
    
        
    \item 
    Another piece of information to consider is whether a pointing gesture has been performed by the HEL. If so, whether the target is an object or a location. This is a categorical feature translated into a vector of size five. The first two elements of the vector are 0 if a pointing gesture hasn't been performed. The first element is 1 if the HEl points to a location, otherwise, if the HEL points to an object, the second element is 1. The last three elements of the vector determine if the location or the object HEL points to matches the ELD's location/object. If the third element of this vector is 1, the HEL points to the same location/object as the ELD, otherwise, if the fourth element is 1, the HEL points to the wrong location/object. If the fifth element is 1, that means the HEL points to the right object type but not the correct specific object. For example, imagine that the ELD asks for a small bowl and the HEL points to a large bowl in response; the vector $(0,1,0,0,1)$ is the indication of the HEL's pointing gesture.
    
    \item Likewise, in the case of H-O actions, we must ascertain whether the action has taken place and whether it was directed toward an object or a location. In addition, we also require the type of H-O (opening or closing a location, touching a location or an object, taking out an object, holding an object). This is a categorical feature, a vector of size ten. The first five elements are interpreted exactly the same as the pointing gesture vector. The second half of the H-O action vector determines which of the five different H-O actions has been performed by one-hot encoding those action types. 

 \item 
 We also add the current HEL’s action and the current HEL’s utterance (DA tag), which are both categorical features represented by vectors of size nine and fourteen respectively. The HEL action classes are categorized as follows: (1) No action, (2) Request $O_T$, (3) Request $L$, (4) Verify $O_T$, (5) Verify $L$, (6) Verify $O$, (7) Acknowledge, (8) Yes, (9) No. The DA classes are categorized as follows: (1) No utterance, (2) Instruct, (3) Acknowledge, (4) Query-w, (5) Query-yn, (6) Reply-w, (7) Reply-y, (8) Reply-n, (9) Check, (10) Explain, (11) Align, (12) State-y, (13) State-n, (14) State. These were the DAs chosen to annotate the "ELDERLY-AT-HOME" corpus~\cite{chen2015roles}.
 
    
 \item 
 In addition to the current HEL info, our model relies on the information from the previous actions from the ELD. Here we use the ELD's action and ELD's DA which are both categorical features represented by vectors of size seven and fourteen respectively. The ELD's action classes are categorized as follows: (1) No action, (2) Give $O_T$, (3) Give $L$, (4) Give $O_T, L$, (5) Acknowledge, (6) Yes, (7) No. The ELD's DA classes are the same as HEL's.
    
\end{itemize}

\subsection{Data Annotation}
The \textit{Find} task data in the ELDERLY-AT-HOME corpus~\cite{chen2015roles} was previously transcribed and annotated for DAs, pointing gestures, and H-O actions. As explained in section \ref{Feature Extraction}, we need to provide the network with additional features for training our user simulator. We performed additional annotations for ELD beliefs of HEL's knowledge of $O_T$, $L$, and $O$, and ELD and HEL actions based on the classes introduced for each feature in the previous section.





ELD's perceptions of HEL's knowledge of $O_T$, $L$, or $O$ were objectively determined based on the heuristic that ELD updates its beliefs whenever HEL demonstrates knowledge or lack thereof of these entities. For instance, ELD assumes that HEL is unaware of the $O_T$ or $L$ until HEL acknowledges ELD's description of it or takes action to confirm it. If HEL selects the wrong $O_T$ or $L$ when ELD has specified a particular one, ELD assumes that HEL is thinking of a different $O_T$ or $L$. For example, if ELD points to a small bowl and says "Get me that bowl," but HEL asks "That bowl?" while pointing to a large bowl, ELD believes that HEL is thinking of a different bowl.


Two annotators labeled the actions of ELD and HEL. As the labeling of action did not follow a strict guideline and was, therefore, more open to interpretation, the inter-annotator agreement was measured for both types of actions using Cohen's kappa. To test the reliability of the annotation, 40 random ELD actions and 40 random HEL actions were chosen from the data set, and both annotators labeled them independently before labeling the remaining actions. The results showed a high level of agreement between the annotators for both ELD actions ($\kappa=1.0$) and HEL actions ($\kappa=0.81$), indicating that the action labels are reliable.


\subsection{Data Augmentation}
\label{Data Augmentation}
The data collected during the \textit{Find} task was suitable for building a strong basis to train the user simulator; however, there were very no instances or few instances in which ELD believed HEL had the wrong $O_T$ or $L$ in mind or did not know the $O_T$. This lack of variation is not surprising, as the interactions between the two humans were relatively straightforward. However, since we propose this user simulator as the main component of the interactive environment for the RL training, and as we expect the HEL agent, trained with reinforcement learning, to make mistakes, we need to augment the data to include examples of missing or infrequent states.

\begin{figure}[h]
\vspace{2mm}
     \centering
     \begin{subfigure}[b]{0.48\textwidth}
         \centering
        \includegraphics[width=\textwidth]{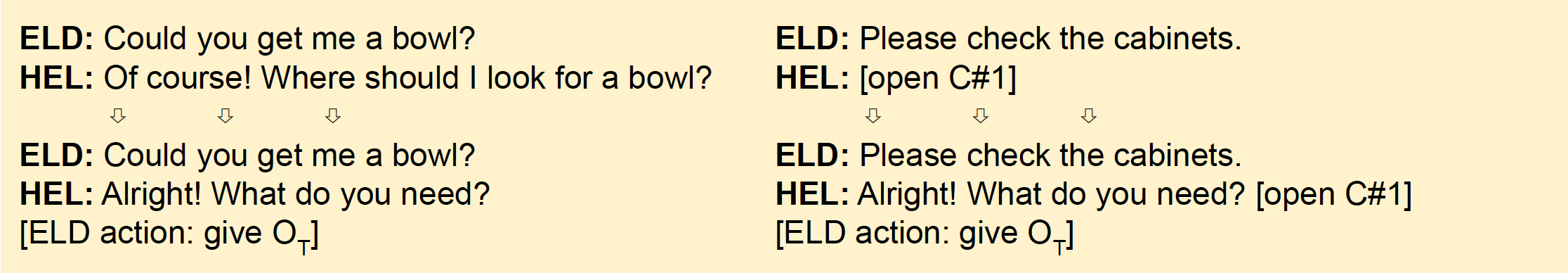}
        \vspace{-6mm}
        \caption{\footnotesize{Generating ``ELD believes HEL does not know $O_T$''}}
        \label{fig:data_aug_0_1_0}
     \end{subfigure}
     \begin{subfigure}[b]{0.48\textwidth}
     \centering
        \includegraphics[width=\textwidth]{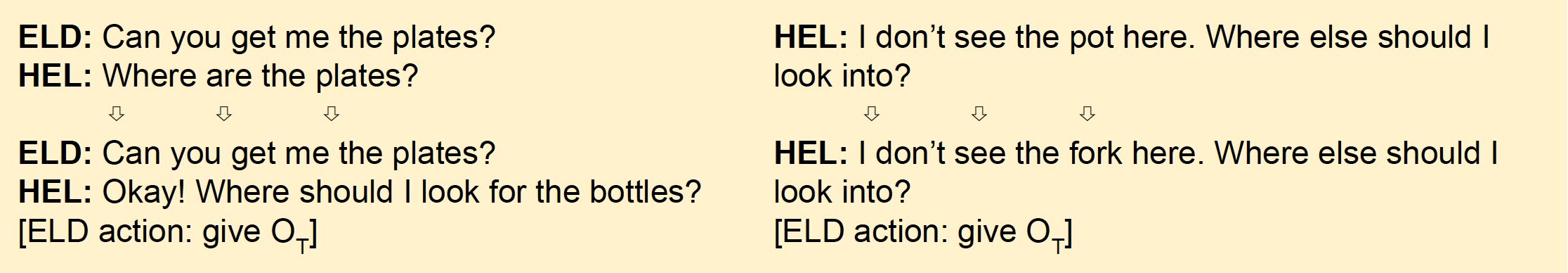}
        \vspace{-6mm}
        \caption{\footnotesize{Generating ``ELD believes HEL has the wrong $O_T$ in mind''}}
        \label{fig:data_aug_2_1_0}
     \end{subfigure}
     \begin{subfigure}[b]{0.48\textwidth}
         \centering
        \includegraphics[width=\textwidth]{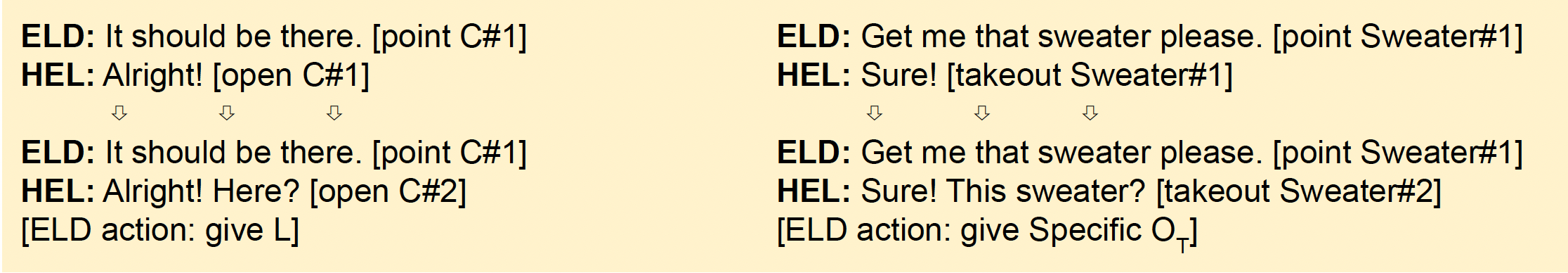}
        \vspace{-6mm}
        \caption{\footnotesize{Generating ``ELD believes HEL is not thinking of the same $L$ or $O$''.}}
        \label{fig:data_aug_1_2_0}
     \end{subfigure}
        \caption{Examples of augmenting the data with unseen or rare states}
        \label{fig:data_aug}
\vspace{-2mm}
\end{figure}

To increase the number of instances in which ELD believes HEL does not know the $O_T$, states (0,0,0) and (0,1,0), we sample instances in which ELD believes HEL knows the $O_T$ and replace HEL's utterance and action with an example of requesting the $O_T$ (see Fig.~\ref{fig:data_aug_0_1_0}). The ELD then gives instructions on $O_T$ again.

To increase the number of instances in which ELD believes HEL has the wrong $O_T$ in mind, states (2, 0, 0) and (2, 1, 0), we sample instances in which HEL mentions the target $O_T$ in their utterance and replace it with an incorrect one (see Fig.~\ref{fig:data_aug_2_1_0}). The ELD then gives instructions on $O_T$ again.

We also increase the number of instances in which ELD believes HEL is not thinking of the same $L$ or $O$, states (1, 2, 0) and (1, 1, 2), by sampling instances in which HEL's utterance or action includes an object or location and replacing it with an incorrect one. In each case, ELD would then give the correct $O_T$ or $L$ again (see Fig.~\ref{fig:data_aug_1_2_0}). These synthetic examples will help the user simulator respond appropriately to mistakes made by the HEL agent.

In our previous study~\cite{shervedani2023multimodal}, we executed the data augmentation only for ELD's output states. However, after carefully examining the data, we observed that only the first nine state combinations mentioned previously are included in the input states whereas all thirteen combinations could be seen in ELD's output states even if they are rare. That being said, state combinations 10 to 13 are meaningful and highly likely to happen during an interaction between the user simulator and the HEL robot agent. Not seeing all possible meaningful inputs during the training, causes the model to be too specific and not able to handle all possible situations it may encounter. It also makes the final accuracy evaluations not accurate because the model has been trained, validated, and tested on inputs that do not offer enough variation. Thus, we augment the data points in such a way that all thirteen state combinations are covered in inputs so that we make our model more accurate and flexible.

To synthesize the input state where in the previous move ELD believed HEL had the wrong $O$ in mind, state (1,1,2), we randomly choose some instances where the input state is (1,1,1), i.e. in the previous move ELD believed HEL had the correct $O$ in mind, and change the ELD's previous move accordingly. For that, ELD would inform the HEL about the object again. So we replace the previous ELD's DA and action with ``Instruct" and ``Give Specific $O_T$" respectively. We should point out that later on, we combined actions "Give Specific $O_T$" and "Give $O_T$" into one single class as during the interaction with the HEL, these two actions convey the same message and only the difference in ELD's state matters when ELD announces either the object type or the specific object during the interaction with HEL.

To synthesize the input states where in the previous move ELD believed HEL had the information about $O_T$ and $L$ but had wrong $O_T$ and/or wrong $L$ in mind, states (1,2,0), (2,1,0), (2,2,0), we randomly sample our data points where in the previous move ELD believed HEL had the right $O_T$ and $L$ in mind and change the ELD's previous move accordingly. For state (1,2,0), ELD would inform the HEL about the location again, so we replace the previous ELD's DA and action with ``Instruct" and ``Give $L$" respectively. Similarly, for the state (2,1,0), ELD would inform the HEL about the object type again, so we replace the previous ELD's DA and action with ``Instruct" and ``Give $O_T$" respectively. Lastly, for the state (2,2,0), ELD would inform the HEL about both object type and location again, so we replace the previous ELD's DA and action with ``Instruct" and ``Give $O_T, L$" respectively.

To synthesize the input states where in the previous move ELD believed HEL had the wrong $O_T$ in mind and didn't have any other information about $L$ and $O$, state (2,0,0), we take random samples where in the previous move ELD believed HEL had the correct $O_T$ in mind and change the previous ELD's DA and action to ``Instruct" and ``Give $O_T$" respectively.

With this data augmentation scheme, we increase our data points from 693 to 1932.

\subsection{Model Architecture and Training}

Our model is a neural network consisting of three fully connected (dense) layers, and a dropout layer (ratio=0.2) to prevent overfitting and improve the ability of the model to generalize better. We utilized the Cross-Entropy loss function and Adam optimizer during training. The training process lasted for a maximum of 100 epochs, but we also evaluated the model's performance on the validation set while training to allow for early stopping. The inputs to the neural network were the features described earlier, while the outputs were the ELD's next state, dialogue act, and action, which were manually annotated in the data. We implemented the model using the PyTorch library~\cite{NEURIPS2019_9015}. 
\section{Evaluation}
\label{Evaluation}

\subsection{Model Evaluation on Data}
The model was trained on $80\%$ of data (a total of 1548 data points). About 10\% of the data (a total of 183 data points) was used for validation purposes during the training to early stop the training before over-fitting happens. The rest of the data (a total number of 201 data points) was used for evaluating the performance of the fully trained model. 

In addition to overall accuracy, we evaluated the model on the classification accuracy of each individual output of the model; i.e. the classification accuracy for (1) the predicted ELD's action; (2) the predicted ELD's DA; (3) the predicted ELD's state, the ELD’s belief of the HEL’s $(O_T, L, O)$.

We should point out that the scores we reported in~\cite{shervedani2023multimodal} were based on testing the BUS model on the original test set, not the augmented test data. However, testing the BUS model on the augmented data would give us similar results. 

Here, we report the accuracy results of BUS and GUS models tested on the original and augmented test data sets. The results are summarized in table \ref{table:1}.

\begin{table}[h]
\vspace{2mm}
\centering
\begin{tabular}{|p{15mm}||c|c|c|c|}
\cline{2-5}
\multicolumn{0}{c||}{}&{\scriptsize Overall Acc.}&{\scriptsize Action Acc.}&{\scriptsize DA Acc.}&{\scriptsize State Acc.}\\
 \hline\hline
{\scriptsize BUS Model,\newline Org. Test Data} &  46.27\% & 52.22\% & 53.73\% &  83.58\%\\
\hline
{\scriptsize BUS Model, Aug. Test Data} &  45.83\% & 56.77\% & 51.04\% &  68.75\%\\
\hline
{\scriptsize GUS Model, Org. Test Data} & 66.67\% & 81.82\% & 72.73 & 93.94\%\\
\hline
{\scriptsize GUS Model, Aug. Test Data} & 70.85\% & 77.89\% & 75.38 & 89.44\%\\
\hline
\end{tabular}
\caption{Classification Accuracy of BUS and GUS Models}
\label{table:1}
\end{table}

\begin{table*}[t]
\vspace{2mm}
\centering
\begin{tabular}{|c||c|c|c|c|c|c|c|c|c|c|c|c|c|c|}
\cline{2-15}
\multicolumn{0}{c||}{}& NoUtt& Inst & Ack & Q-w & Q-yn & R-w & R-y & R-n & Chk & Exp & Algn & St-y & St-n & St\\
 \hline\hline
No Utt & \textbf{67.79\%} & 0\% & 0\% & 0\% & 0\% & 1.69\% & 16.95\% & 0\%& 0\% & 0\%& 0\% & 6.78\%& 0\% & 0\%\\
\hline
Inst & 12\% & \textbf{46\%} & 0\% & 0\%& 0\% & 12\% & 0\% & \textbf{30\%} & 0\% & 0\%& 0\% & 0\% & 9.09\% & 0\%\\
\hline
Ack & 0\% & 0\% & \textbf{66.67\% }& 0\% & 0\% & 0\% & 0\% & 0\% & 0\% & 0\% & 0\% & \textbf{33.33\%} & 0\% & 0\%\\
\hline
Q-w & $-$ & $-$ & $-$ & $-$ & $-$ & $-$ & $-$ & $-$ & $-$ & $-$ & $-$ & $-$ & $-$ & $-$\\
\hline
Q-yn & $-$ & $-$ & $-$ & $-$ & $-$ & $-$ & $-$ & $-$ & $-$ & $-$ & $-$ & $-$ & $-$ & $-$\\
\hline
R-w & 0\% & 2.32\% & 0\% & 0\% & 0\% & \textbf{93.03\%} & 4.65\% & 0\% & 0\% & 0\% & 0\% & 0\% & 0\% & 0\%\\
\hline
R-y & 0\% & 0\% & 0\% & 0\% & 0\% & 0\% & \textbf{100\%} & 0\% & 0\% & 0\% & 0\% & 0\% & 0\% & 0\%\\
\hline
R-n & 0\% & 20\% & 0\% & 0\% & 0\% & 0\% & 0\% & \textbf{80\%} & 0\% & 0\% & 0\% & 0\% & 0\% & 0\%\\
\hline
Chk & $-$ & $-$ & $-$ & $-$ & $-$ & $-$ & $-$ & $-$ & $-$ & $-$ & $-$ & $-$ & $-$ & $-$\\
\hline
Exp & $-$ & $-$ & $-$ & $-$ & $-$ & $-$ & $-$ & $-$ & $-$ & $-$ & $-$ & $-$ & $-$ & $-$\\
\hline
Algn & $-$ & $-$ & $-$ & $-$ & $-$ & $-$ & $-$ & $-$ & $-$ & $-$ & $-$ & $-$ & $-$ & $-$\\
\hline
St-y & 20\% & 0\% & 0\% & 0\% & 0\% & 0\% & 0\% & 0\% & 0\% & 0\% & 0\% & \textbf{80\%} & 0\% & 0\%\\
\hline
St-n &0\% & 0\% & 0\% & 0\% & 0\% & 0\% & 0\% & 0\% & 0\% & 0\% & 0\% & 0\% & \textbf{100\%} & 0\%\\
\hline
St & $-$ & $-$ & $-$ & $-$ & $-$ & $-$ & $-$ & $-$ & $-$ & $-$ & $-$ & $-$ & $-$ & $-$\\
\hline
\end{tabular}
\caption{DA Classification Confusion Matrix. Cells with $-$ indicate that there were no such DA labels in the evaluation data set and the model did not predict any of such DAs.}
\label{table:2}
\end{table*}

\begin{table*}[t]
\centering
\begin{tabular}{|c||c|c|c|c|c|c|c|}
\cline{2-8}
\multicolumn{0}{c||}{}& No Act& Give $O_T$ & Give $L$ & Give $O_T, L$ & Ack & Yes & No\\
 \hline\hline
No Act & \textbf{74.08\%} & 0\% & 0\% & 0\% & 0\% & 25.92\% & 0\%\\
\hline
Give $O_T$ & 8.82\% & \textbf{69.12\%} & 0\% & 0\% & 0\% & 22.06\% & 0\%\\
\hline
Give $L$ & 10\% & 0\% & \textbf{83.33\%} & 0\% & 0\% & 6.67\% & 0\%\\
\hline
Give $O_T, L$ & $-$ & $-$ & $-$ & $-$ & $-$ & $-$ & $-$\\
\hline
Ack & 0\% & 0\% & 0\% & 0\% & \textbf{66.67\%} & 33.33\% & 0\%\\
\hline
Yes & 0\% & 3.12\% & 0\% & 0\% & 0\% & \textbf{96.88\%} & 0\%\\
\hline
No & 0\% & 11.11\% & 0\% & 0\% & 0\% & 0\% & \textbf{88.89\%}\\
\hline
\end{tabular}
\caption{Action Classification Confusion Matrix. Cells with $-$ indicate that there were no such Action labels in the evaluation data set and the model did not predict any of such Actions.}
\label{table:3}
\end{table*}

The results in table \ref{table:1} show a great improvement from the BUS model to the GUS model. This significant improvement results from the changes we made to our feature extraction, data augmentation, and the model architecture itself that can be summarized as (1) changing the ELD's state representation as explained in detail in section \ref{Feature Extraction}, (2) Giving information about ELD's previous DA and action as input to the model, (3) Augmenting ELD's input states as explained in detail in section \ref{Data Augmentation}, (4) Adding layers to the network architecture and removing nonlinear ReLU activation function from the network
, (5) Not executing sample re-weighting since the class imbalance issue is already resolved by proper data augmentation and we don't want to overdo re-weighting because of the limited amount of data we have. Considering the small amount of data, re-weighting actually affects the performance of our model adversely.

To further analyze the evaluation results reported in table \ref{table:1}, we investigate the corresponding confusion matrices. Before moving forward with the confusion matrices, one should remember that as explained in section \ref{Feature Extraction}, our trained model is supposed to also decide whether or not the ELD will take a move. If that's the case when the HEL is taking two consecutive moves, the model would output \emph{0} and \emph{None} as the predicted ELD's action and DA respectively. 

The DA confusion matrix in table \ref{table:2} shows that the main confusion is due to wrongly classified ``Instruct" DAs as ``Reply-n". One justification for this confusion is the limited number of data, and on top of that is the imbalanced classes. However, going over the annotated data, we see that most of the time when the ELD responds to a verification question or $O_T/L$ query (``Query-yn/Cehck" and ``Query-w" DAs respectively), s/he carries on with giving further instructions. That means in many cases where ELD's utterances are labeled as ``Reply-n", they could also be interpreted as ``Instruct" and eventually convey the same intent to the HEL. For instance, the HEL verifies the $O_T$ by asking the ELD ``Did you say a pot?", and the ELD replies with ``No, get me a bowl.". The ELD's DA could be labeled as either ``Reply-n" or "Instruct", but what matters here is that the HEL receives the same intent from ELD.

Going through the confusion matrix in table \ref{table:2}, we observe that the rest of the DA classes are either classified very well or confused with classes that don't influence the overall outcome of the network. For instance, the DA ``Acknowledge" is correctly classified in 66.67\% of the cases and has been classified as ``State-y" in 33.33\% of the cases. However, this misclassification doesn't affect the message that HEL receives.

Analysis of the action confusion (the results presented in table \ref{table:3}) shows that 22.06\% of the "Give $O_T$" samples are wrongly classified as ``Yes". This is again due to the limited number of data points available as well as imbalanced classes. However, analogous to our explanation above, this misclassification could be because of those cases where the ELD responds to a verification question and continues by giving instructions. For example, the HEL verifies the $L$ by asking the ELD ``Did you say that cabinet?", and the ELD replies with ``Yes, get me the silverware.". The ELD's action could be labeled as either "Yes" or "Give $O_T$". In either case, the HEL would receive the same message.

We also observe that 33.33\% of the ``Acknowledge" classes are misclassified as ``Yes" actions. Again, because these two actions are inherently very similar, this misclassification doesn't affect the message that HEL receives.

In summary, many wrong DA and action classifications are due to the fact that distinguishing DA classes like ``Instruct", ``Reply-y", ``Reply-n", ``State-y", ``State-n", and action classes such as ``Give $O_T$", ``Give $L$", ``Acknowledge", ``Yes", and ``No" would be very difficult. This happens because in our available data, most of the cases where the ELD responds to a ``Query-yn" question, start with saying yes, no, acknowledging, and then guiding the HEL towards a location and/or giving information about the object.


\subsection{Model Evaluation on HBATN}
\label{HBATN_evaluation}

\begin{table*}[t]
\vspace{2mm}
\centering
\begin{tabular}{|c||c|c|c|}
\cline{2-4}
\multicolumn{0}{c||}{}& Input & HBATN Output & GUS Output\\
 \hline\hline
Establish($O_T$) & $(0,0,Inst)/(0,0,Qw)$ & $(1,*,Inst)/(1,*,Rw)$ & $(1,0,Rw)$\\
\hline
Verify($O_T$) & $(1,0,Chk)/(1,0,Qyn)$ & $(*,0,Ry)/(*,0,Rn)/(1,0,Inst)/(1,0,Rw)$ & $(1,0,Ry)/(1,1,Rn)/(1,0,Inst)/(1,*,Rw)$\\
\hline
Specify($O_T$) & $(*,0,Qw)$ & $(*,0,Inst)/(*,0,Rw)$ & $(*,0,Rw)$\\
\hline
\end{tabular}
\caption{GUS performance evaluations for Det$(O_T)$ subtask. $(*)$ represents 0 or 1.}
\label{table:4}
\end{table*}

\begin{table*}[t]
\vspace{2mm}
\centering
\begin{tabular}{|c||p{30mm}|p{55mm}|c|}
\cline{2-4}
\multicolumn{0}{c||}{}& Input & HBATN Output & GUS Output\\
 \hline\hline
Establish($L$) & $(*,0,Inst)/(*,0,Qw)$ & $(*,*,Inst)/(*,*,Rw)$ & $(*,*,Inst)/(*,*,Rw)/(*,*,Ry)/(*,*,Rn)$\\
\hline
Verify($L$) & $(0,0,-)/(0,*,Chk)/$ \newline$(0,*,Qyn)$ & $(0,*,Ry)/(0,*,Rn)/(*,*,-)/$ \newline$(*,*,Rw)/(*,*,Inst)$ & $(0,*,Sty)/(0,*,Ry)/(0,*,Rn)$\\
\hline
Specify($L$) & $(*,1,Qw)$ & $(*,*,Inst)/(*,*,Rw)$ & $(0,*,Inst)/(0,*,Rn)/(*,*,-)$\\
\hline
\end{tabular}
\caption{GUS performance evaluations for Det$(L)$ subtask. $(*)$ represents 0 or 1.}
\label{table:5}
\end{table*}

\begin{table*}[t]
\vspace{2mm}
\centering
\begin{tabular}{|c||p{30mm}|c|c|}
\cline{2-4}
\multicolumn{0}{c||}{}& Input & HBATN Output & GUS Output\\
 \hline\hline
Specify($O_T$) & $(*,0,Qw)$ & $(*,0,Inst)/(*,0,Rw)$ & $(*,*,Inst)/(*,*,Rw)/(*,0,Ry)/(*,*,Rn)$\\
\hline
Verify($O$) & $(*,0,-)/(*,0,Chk)/$ \newline$(*,0,Qyn)/(*,*,St)$ & $(*,0,Ry)/(*,0,Rn)/(*,0,Rw)/(*,0,Inst)$ & $(*,0,Ry)/(*,0,Rn)/(*,*,Inst)$\\
\hline
Finish($L$) & $(*,*,Sty)(*,*,St)/$ \newline$(*,*,Stn)$ & $(0,0,Ack)$ & $(0,0,Ry)/(0,0,Sty)$\\
\hline
\end{tabular}
\caption{GUS performance evaluations for Det$(O)$ subtask. $(*)$ represents 0 or 1.}
\label{table:6}
\end{table*}

To further investigate how realistic our model performance is, we compare its performance to our previously developed HBATN model. It is important whether or not the GUS model acts similarly to HBATNs because our HBATN is carefully hand-crafted by human annotators who based their insights in the data. For the purpose of this comparison, we would need to have the response of our trained model to variant inputs.

To generate different meaningful inputs for our GUS model, we employ an automatic approach. Before moving to explain the approach itself, we need to lay some ground rules as follows:
\begin{itemize}

\item For the ELD's state representation, ELD's belief of HEL's knowledge of $O_T$ cannot change from 0 to 1 or 2 before ELD utters the $O_T$.
\item ELD's belief of HEL's knowledge of $L$ cannot change from 0 to 1 or 2 before ELD utters the $L$.
\item ELD's belief of HEL's knowledge of $O$ cannot change from 0 to 1 or 2 before ELD's belief of HEL's knowledge of $O_T$ turns to 1.
\item ELD's belief of HEL's knowledge of $O$ can change from 0 to 1 or 2 before ELD's belief of HEL's knowledge of $L$ turns to 1.
\item HEL cannot verify $O_T$ and $L$ before ELD announces them.
\item HEL only performs pointing and H-O actions for $O_T/L/O$ verifications.
\end{itemize}

The ground rules associated with the ELD's state representation would give us thirteen distinct meaningful combinations which we explained in detail in section \ref{Feature Extraction}. After combining those states with other inputs, by applying the rest of the ground rules to all combinations, we generate all meaningful inputs automatically.

Subsequently, we put our GUS model in different states we previously had extracted from our HBATN~\cite{monaikul2020role} by applying different inputs to the model and finally we compare the GUS model outputs to those of HBATN.

The results of comparing our GUS model to the HBATN are summarized in tables \ref{table:4}, \ref{table:5}, \ref{table:6}. In the $(a,b,c)$ tuple which represents the input or the output, $c$ stands for the HEL DA (input) or ELD DA (output), $a$ and $b$ also determine whether or not the $O_T$ and $L$ have been uttered respectively. Each input/output also includes ELD's previous state, ``pointing/H-O" actions, ELD's previous DA and action, and HEL's DA and action features. We omitted these parameters in our table representations for simplicity. However, we carefully mapped our automatically generated inputs to different HBATN states for these comparisons.

Our comparisons illustrate that the GUS outputs greatly match the HBATNs. There are only a few minor differences that don't affect the interaction between the ELD and the HEL and their intents. For example, our GUS model confuses ``Instruct" and ``Reply-w" DAs in some cases such as in table \ref{table:4}, the primitive subtask ``Establish$(O_T)$" where it only outputs ``Reply-w" for all ``Instruct/Reply-w"-labeled outputs. For the HEL it doesn't matter if the utterance is labeled as either one because both DAs transfer the same message.

In some other instances, such as in table \ref{table:5}, the primitive subtask ``Specify$(O_T)$", our GUS model outputs ``Instruct/Reply-y/Reply-n" DAs for ``Instruct"-labeled outputs. This again is not a fatal error because the action the GUS model outputs as the ELD's action would provide instructions about the $O_T$ or $L$. In table \ref{table:6}, for the primitive subtask ``Finish$(L)$", the GUS model confuses ``Acknowledge" DA with ``Reply-y/State-y". This is also negligible due to the similar inherent that these DAs carry. Nevertheless, this is the final action in the interaction and wouldn't affect the interaction at all.

We should also point out that our GUS model differs from the HBATNs in some cases where the ELD in HBATNs utters or doesn't utter the $O_T$ or $L$. In a few cases, the GUS model utters $O_T$ or $L$ when the ELD in HBATNs doesn't or vice versa. This minor error is due to applying machine learning approaches to build an artificial intelligence agent as our user simulator. Although these minor errors could make the interaction longer or unsuccessful in very few cases, our GUS model would still be superior to the hand-crafted framework. The reader may refer to our previous study for an elaborated explanation of the interaction between our user simulator and Reinforcement Learning HEL agent ~\cite{shervedani2023multimodal} which demonstrates that our user simulator can be successfully used in practice for the RL training of assistive robots. 

\section{Conclusion}

\label{conclusion}


This paper introduces a new user simulator that uses a neural network to provide a diverse interactive environment for training RL agents in collaborative tasks with multiple modes of communication. The user simulator is based on the ``Find'' task demonstrations from the ELDERLY-AT-HOME interaction corpus. It is capable of interpreting and responding to multiple modalities, such as language, pointing gestures, and haptic-ostensive actions. The primary contribution of the study is the creation of a user simulator that can provide a multimodal interactive environment for RL training, which has not been done before. The secondary contribution is a novel multimodal data augmentation approach that effectively overcomes the challenges of using limited and sparse human demonstrations to develop intelligent data-driven agents. The developed simulator has been evaluated both to test its ability to represent the training data and to compare its performance with the manually crafted HBATN models. The results show a very good performance of the simulator and suggest that it can be successfully used in the RL training of domestic assistive robots.

\bibliographystyle{IEEEtran}
\bibliography{references}

\end{document}